\documentclass{article} 
\usepackage{iclr2021_conference,times}

\usepackage{enumitem}

\usepackage{amsmath,amsfonts,bm}

\def\eqref#1{equation~\ref{#1}}

\def\1{\bm{1}}

\DeclareMathAlphabet{\mathsfit}{\encodingdefault}{\sfdefault}{m}{sl}
\SetMathAlphabet{\mathsfit}{bold}{\encodingdefault}{\sfdefault}{bx}{n}

\usepackage{amsthm}
\usepackage{graphicx}
\usepackage{amsfonts}
\usepackage{algorithm}
\usepackage[noend]{algpseudocode}
\usepackage[english]{babel}

\usepackage{hyperref}
\usepackage{url}

\title{Learning compositional structures for deep learning: why routing-by-agreement is necessary}

\author{Sai Raam Venkataraman, Ankit Anand, S. Balasubramanian, \& R. Raghunatha Sarma \\
	    Department of Mathematics and Computer Science, Sri Sathya Sai Institute of Higher Learning \\
}

\begin{document}

\maketitle

\begin{abstract}
A formal description of the compositionality of neural networks is associated directly with the formal grammar-structure of the objects it seeks to represent. This formal grammar-structure specifies the kind of components that make up an object, and also the configurations they are allowed to be in. In other words, objects can be described as a parse-tree of its components - a structure that can be seen as a candidate for building connection-patterns among neurons in neural networks. We present a formal grammar description of convolutional neural networks and capsule networks that shows how capsule networks can enforce such parse-tree structures, while CNNs do not. Specifically, we show that the entropy of routing coefficients in the dynamic routing algorithm controls this ability. Thus, we introduce the entropy of routing weights as a loss function for better compositionality among capsules. We show by experiments, on data with a compositional structure, that the use of this loss enables capsule networks to better detect changes in compositionality.  Our experiments show that as the entropy of the routing weights increases, the ability to detect changes in compositionality reduces. We see that, without routing, capsule networks perform similar to convolutional neural networks in that both these models perform badly at detecting changes in compositionality. Our results indicate that routing is an important part of capsule networks - effectively answering recent work that has questioned its necessity. We also, by experiments on SmallNORB, 
CIFAR-10 and FashionMNIST, show that this loss keeps the accuracy of capsule network models  comparable to models that do not use it .  
\end{abstract}

\section{Introduction}
The compositionality of visual objects can be described as being a parse-tree of objects, where objects at a level of this tree combine to form the object they are connected to at the higher-level. The types of objects, and their allowed configurations for combination can be described by category-specific rules. Therefore, the compositionality of an object can also be described by a spatial grammar, such that the derivation-rules of this grammar describe the parse-tree of the object.
\paragraph{}
This presents a candidate for the kind of connection-patterns among activations we wish to achieve in compositional neural networks - the representation of an object must be given by a tree-like set of activations across multiple layers. Each connection between activations of two layers can be thought of as representing a part-whole relationship between the objects they represent, and therefore a learned derivation-rule of some spatial grammar. In a setting with several compositionalities among inputs, this tree-like structure means that an activation in a layer of a neural network can be strongly connected to only one activation of a deeper layer.
\paragraph{}
What would be necessary, for such a structure, is a mechanism that decides what importance a derivation-rule is assigned. In our work, we identify that no such mechanism exists for convolutional neural networks (CNNs), while capsule networks \citep{sabour2017dynamic}, \citep{hinton2018matrix}, \citep{sairaam2020building} can decide the importances of derivation-rules by their process of building deeper activations.
\paragraph{}
We briefly discuss this procedure. Deeper capsules are built from a consensus-based aggregation of predictions made for them by shallower capsules. By having capsules activated only when predictions for them are in agreement, this procedure allows for a checking of whether the shallower capsules share a common viewpoint and can  form the corresponding object. This process is termed routing.
\paragraph{}
A class of routing algorithms, for example, dynamic routing  \citep{sabour2017dynamic} aggregates predictions by a weighted-sum, each weight denoting the extent of the consensus of a prediction for a deeper capsule relative to the consensus of the predictions of the corresponding capsule for other deeper capsules. For a parse-tree structure, only one routing-weight for a shallower capsule must be strong. We identify that a low entropy of routing weights for each capsule can improve the compositional structure of capsule networks.
\paragraph{}
However, this does not happen in the dynamic routing where we show that the entropy of routing weights is high. To remedy this, we minimize entropy of routing weights, and subsequently demonstrate that we are able to achieve better compositional representations. A measure of this can be seen in the ability to detect changes in compositional structures. Figure \ref{changedcompostionality} shows an example of such a change. We can measure a network's compositionality by observing its behaviour on such data. We show that CNNs do not detect differences between such data, while capsule networks can. We show that capsule networks with a low entropy of routing weights detect such differences more than capsule networks with a higher entropy of routing weights. This work, to our knowledge, is the first study of the compositional structure of capsule networks and routing.
\paragraph{}
We also show that entropy-regularised capsule networks achieve similar performance to unregularised networks on classification tasks on SmallNORB, 
CIFAR-10 and FashionMNIST. We list our contributions below:
\begin{figure}\label{changedcompostionality}
	\centering
	\caption{Example of a change in compositionality in face images. The image on the left shows the compositionality of a face. The image on the right has the same parts, but not the same relationships between them. Images are from \citep{facevsnonface})}.
	\includegraphics[scale=0.3]{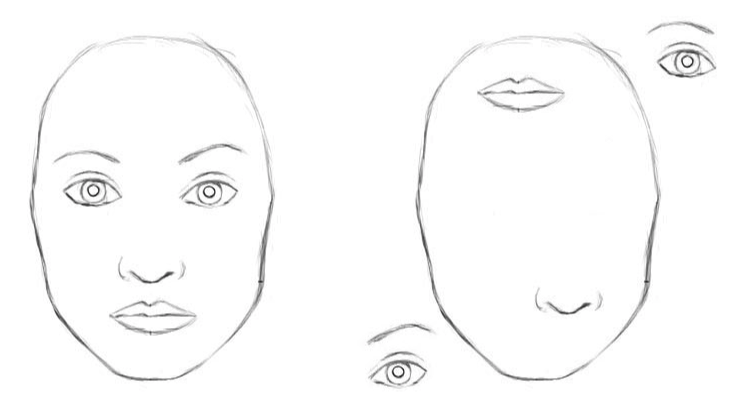}
\end{figure}
\begin{itemize}
	\item[1.] A formal language description of CNNs and capsule networks to show their difference in compositionalities. (see section \ref{FL})
	\item[2.] Using entropy of the routing weights as a loss function for better compositionality. (see section \ref{ENT})
	\item[3.] Experiments on capsule networks that show that entropy-regularised capsule networks are able to detect changes in compositional structures. Unregularised capsule networks do not show the same extent of this ability, and CNNs do not show it. (see section \ref{EXP1})
	\item[4.] Experiments on CIFAR10 and FashionMNIST that show entropy-regularised capsule networks have similar accuracy to other capsule networks. (see section \ref{EXP2})
\end{itemize}

\section{Related work}
Grammar-models are widely used in computer vision, for example, \citep{zhu2007stochastic}, \citep{tu2015stochastic}, \citep{lin2009stochastic}, \citep{zhao2011image}. These models learn a grammar-structure directly for representing the compositionality of objects, and use the learned grammar in computer vision tasks. There are several issues with directly implementing grammar models such as in \citep{zhu2007stochastic}; for example, the need for annotations of parts. Further problems are the difficulty in scaling to deeper structures and relatively lower performance - as discussed in \citep{tang2017towards}. One means of having high-performing grammatical models is to use grammar-concepts in deep learning. Some work in this direction can be seen in \cite{li2019aognets} and \cite{tang2017towards}. However, these models are not a parse-tree representation, and do not incorporate any notion of choosing between spatial relationships among objects, which is what we wish to do.
\paragraph{}
An alternative approach that is not as grounded in formal-language theory, but attempts to build tree-like representations for data is the capsule network model. Several capsule network models exist that detail various routing algorithms and architectures: \citep{sabour2017dynamic}, \cite{hinton2018matrix}, \citep{wang2018optimization}, \citep{ahmed2019star}, \citep{rajasegaran2019deepcaps}, \citep{sairaam2020building}, and \cite{choi2019attention}. However, these and many other models that we have not mentioned do not address the issue of analysing part-whole relationships. A recent work by \citep{peer2018increasing} works on building a new routing algorithm that reduces the entropy of routing-weights. However, their paper does not test the compositionality of capsule networks - and, as we show, does not reduce the entropy to a small value. \citep{kissner2019neural} proposes a  grammar-structure based on units termed capsules. This work does not, however, propose a means of studying the compositionality of models, and does not learn compositional structures from unannotated inputs.  
\section{Formal Language Description of CNNs and Capsule Networks}\label{FL}
In this section, we formally describe a grammar-structure for CNN-models and capsule networks. We begin by doing so for CNNs. Before we present a grammar-description of CNN-models, we first present definitions for the layers that our grammar describes.
\subsection{CNN-models}
\paragraph{}
We consider CNN-models as functions on groups so that our work extends to group-equivariant CNNs \citep{cohen2016group}. The following definitions follow from the same work.
\begin{description}\label{cnnlayer}
\item[\textbf{Correlation:}] Consider a group ($G,.$), functions $f$ : $G \times (\mathbb{N}\cup\{0\})$ $\rightarrow \mathbb{R}$ and $\Psi$ : $G \times (\mathbb{N}\cup\{0\})$ $\times (\mathbb{N}\cup\{0\})$ $\rightarrow \mathbb{R}$. Then, the correlation operator is given by ($f\star \Psi$)$(g,i)$ = $\Sigma_{h\in G}^{}$ $\Sigma_{k=0}^{d-1}$ $(f(h,k)\Psi(g^{-1}.h,k,i))$ where d is the number of inchannels. 
\item[\textbf{Activation function:}] Consider a group ($G,.$), and a function $f$ : $G \times (\mathbb{N}\cup\{0\})$ $\rightarrow \mathbb{R}$. Let $v$ : $\mathbb{R}$ $\rightarrow$ $\mathbb{R}$. Then $C_{v}f(g,n)$ = $v(f(g,n))$ represents the point-wise activation function.
\item[\textbf{Max-pooling:}] As in \citep{cohen2016group}, we split max-pooling into two steps, pooling and subsampling. Consider a group ($G,.$). Let $f$ : $G$ $\times \mathbb{N}\cup\{0\}$ $\rightarrow$ $\mathbb{R}$. Further, let $U$ be a subset of $G$. Then, the pooling step is defined by $MaxPool f(g,i)$ = $max_{k\in g.U}f(k)$. The subsampling step, which is used in the case of strides, is given by subsampling the maxpooled function on a subgroup of $G$. 
\end{description}
\paragraph{}
We consider a layer of such a model to be a correlation-layer followed by an activation function, or a max-pooling layer, or layer resulting in a weighted sum of two layers. The grammar for such a model can be described as follows.
\paragraph{}
We define a grammar defined on a group ($G, .$) as the tuple ($\Sigma, N, S, R, f$), where:
\begin{description}
	\item[\textbf{$\Sigma$}] is the set of terminals. These denote patterns that do not have parts.
	\item[\textbf{$N$}] is the set of non-terminals. These denote patterns in the intermediate layers of the compositionality.
	\item[\textbf{$S$}] is the start symbol. This denotes a complete pattern.
	\item[\textbf{$R$}] is the set of derivation-rules. Each derivation-rule can be one of two types:
	\begin{description}
		\item[\textbf{1}] $X_{1} \vert X_{2} ... \vert X_{n}$ $\rightarrow$ $C$, where $X_{1}, X_{2}, ..., X_{n}$ $\in$ $N$, and $C$ is a sequence of terminals and non-terminals.
		\item[\textbf{2}] $X$ $\rightarrow$ $C_{1}C_{2}...C_{n}$, where $C_{1}, C_{2}, ..., C_{n}$ $\in$ $\Sigma\cup N$, and $X$ $\in$ $\{C_{1}, C_{2}, ..., C_{n}\}$.
	\end{description}
    \item[\textbf{$f$}] is a function $f$ : $N$ $\times$ $G$ $\times$ $\mathbb{N}\cup\{0\}$ $\rightarrow$ $\mathbb{R}$. $f$ denotes the activation of an instance of a non-terminal.
\end{description}
\paragraph{}
CNNs are an example of this grammar. The input is treated as a sequence of terminals. Each activation is treated as an instance of a non-terminal. Further, each process of obtaining a deeper activation is treated as an application of a derivation-rule. Two types of rules exist. The first type can be described as an enumeration of alternate derivations from a set of non-terminals. This corresponds to the non-pooling layers of the CNN, where a local pool of activations is combined using multiple convolutions to obtain more than one activation, in general.  
\paragraph{}
The second is a subsampling rule, that chooses one non-terminal or terminal from a set of terminals and non-terminals. This represents the max-pooling layers.
\paragraph{}
The activation function in the grammar gives the actual value of the activation associated with each instance of a non-terminal.
\subsection{Capsule networks}
Capsule networks are a recent family of neural networks. Each activation of a capsule network is a vector that represents the pose of an instance of an object of some type. Deeper capsules are built through routing by combining predictions made by shallower capsules for them in a consensus-based manner. Routing aims to capture the level of agreement in viewpoints among components.
\paragraph{}
Several capsule models exist, with multiple prediction strategies and routing algorithms \citep{sabour2017dynamic}, \citep{sairaam2020building}, \citep{rajasegaran2019deepcaps}, \citep{hinton2018matrix}. In our work, we use SOVNET model Venkataraman et al. (2020) with dynamic routing.
\paragraph{}
The SOVNET model associates each capsule-type with a neural network defined on a group ($G,.$). Shallower capsules use this neural network to make predictions for deeper capsules. The procedure for using dynamic routing while using convolutional predictions is given in Algorithm \ref{capsulerouting}.
\begin{algorithm}
	\caption{The dynamic routing algorithm for SOVNET}\label{capsulerouting}
	
	\textbf{Input}:  $\{f_{i}^{l}|i\in\{0,...,N_{l}-1\}, f_{i}^{l}:G\rightarrow \mathbb{R}^{d^{l}}\}$ where $N_l$ is number of capsules in layer $l$\\
	\textbf{Output}: $\{f_{j}^{l+1}|j\in\{0,...,N_{l+1}-1\}, f_{j}^{l+1}:G\rightarrow \mathbb{R}^{d^{l+1}}\}$\\
	\textbf{Trainable Functions}:
	$(\Psi_{j}^{l+1},\star)$, $0 \leq j \leq N_{l+1} -1$, - a set of $d^{l+1}$ group-equivariant convolutional filters (per capsule-type) that use the group-equivariant correlation operator $\star$\\
	\hspace*{\algorithmicindent} $S_{ijp}^{l+1}(g)$ = $(f_{i}^{l} \star \Psi_{j}^{l+1,p})(g)$ = $\sum_{h\in G}\sum_{k=0}^{d^{l}-1}f_{ik}^{l}(h)\Psi_{k}^{l+1,p}(g^{-1} \circ h)$; $p \in \{0,...,d^{l+1}-1\}$\newline
	\hspace*{\algorithmicindent} $b_{ij}^{l+1}(g)$ = 0 $\forall 0 \leq i \leq N_{l}-1$, $\forall 0 \leq j \leq N_{l+1}-1$, $\forall g \in G$
	\begin{algorithmic}
	\For{iter in ITER}
	    \State $c_{ij}^{l+1}(g)$ $\gets$ $\frac{exp(b_{ij}^{l+1}(g))}{\Sigma_{k=0}^{N_{l+1}-1}exp(b_{ik}^{l+1}(g))}$
	    \State $f_{j}^{l+1}(g)$ = $\sum_{i=0}^{N_{l}-1} c_{ij}^{l+1}(g)S_{ij}^{l+1}(g)$ $\forall$ $ 0  \leq j \leq N_{l+1} - 1$, $\forall g \in G$
	    \State $f_{j}^{l+1}(g)$ = $Squash(f_{j}^{l+1}(g))$ =  $\frac{\Vert f_{j}^{l+1}(g) \Vert}{1 +\Vert f_{j}^{l+1}(g) \Vert^{2}}$ $f_{j}^{l+1}(g)$; $\forall$ $0 \leq j \leq N_{l}-1$, $\forall$ $g$ $\in$ $G$
	    \State $b_{ij}^{g}$ = $b_{ij}^{l+1}(g)$ + $S_{ij}^{l+1}(g).f_{j}^{l+1}(g)$, $\forall g \in G$, $\forall 0 \leq i \leq N_{l}-1$, $\forall 0 \leq j \leq N_{l+1}-1$
	\EndFor
	\end{algorithmic}	
\end{algorithm}
\paragraph{}
We now describe this capsule network model in terms of a grammar. Consider the grammar defined on a group ($G,.$), and given by the tuple ($\Sigma, N, S, R, f$), where
\begin{description}
	\item[\textbf{$\Sigma$}] is the set of terminals.
	\item[\textbf{$N$}] is the set of non-terminals.
	\item[\textbf{$S$}] is the start symbol, and represents a complete pattern.
	\item[\textbf{$R$}] is the set of derivation-rules. Derivation-rules are  of two types.
	\begin{description}
		\item[\textbf{OR-rule}] $X_{1} \vert X_{2} ... \vert X_{n}$ $\rightarrow$ $C$, where $X_{1}, X_{2}, ..., X_{n}$ $\in$ $N$, and $C$ is a sequence of terminals and non-terminals. Further, there exists a likelihood-function $p$ that maps each derivation $X_{i}$ $\rightarrow$ $C$ to a number between 0 and 1, such that $\Sigma_{i=0}^{n}p(X_{i})$ = 1.
		We term $X_{1}, X_{2}, ..., X_{n}$ as OR-nodes.
		\item[\textbf{AND-rule}] $X$ $\rightarrow$ $C_{1}C_{2} ... C_{I}$, where $X$ $\in$ $N$, and $C_{1}, C_{2}, ..., C_{I}$ are OR-nodes. $X$ is termed an AND-node. 
 	\end{description}
    \item[\textbf{$f$}] is a function $f$ : $N \times G \times \mathbb{N}\cup\{0\}$ $\rightarrow$ $\mathbb{R}^{d}$. $f$ represents the activation of an instance of a non-terminal.
\end{description}
\paragraph{}
Capsule networks as given in Algorithm \ref{capsulerouting} are an example of this grammar. The terminals denote the input. Each prediction made by a capsule for a deeper capsule is an OR-rule. The process of assigning routing-weights is the likelihood function. The AND-rule combines predictions to form the deeper capsules.    
\section{Entropy among derivations}\label{ENT}
In order to have a parse-tree structure of active derivations, the likelihood function for an OR-rule must have only one large value. This corresponds to a low entropy. In the case where the entropy is large, there is little difference between CNNs and capsule networks in terms of compositionality. This is because in such a case, there is little difference in the importance among alternate derivations. 
\paragraph{}
The entropy of the routing-weights of capsule networks with dynamic routing, however, are not always low \citep{peer2018increasing}. In which case, to have a compositional representation, the entropy must be reduced. We propose a loss function to this end.
\paragraph{}
We use the sum of the mean entropies of the routing-weights $c_{ij}^{l}(g)$ for each layer of the network. Thus, in addition to a loss function for the task, we also use a loss for compositionality. This is an alternate approach to the work of \cite{peer2018increasing}, where a routing algorithm, termed scaled-distance-agreement routing, was used to reduce the entropy of routing-weights.
\section{Experiments}
We conduct two sets of experiments to test the compositionality and accuracy of entropy-regularised capsule networks. The first experiment involves training models on data with a compositional structure and testing it on data with a different compositionality. The second experiment involves checking the accuracy of models on SmallNORB, 
CIFAR10 and FashionMNIST. We describe each of these experiments in more detail in the following sections.
\subsection{Detecting changes in compositional structures}\label{EXP1}
Since we are interested in testing the ability of models to detect changes in compositional structure, we constructed a dataset where this is possible. We used the task of determining if an input is a face-image or not as the learning-task. We randomly selected 50,000 images of the MORPH dataset - each image frontalised and aligned - for the train-set of faces, and used the train-set - of 50,000 images - of the Animal-10N dataset for non-face images. The images of the MORPH dataset\citep{rawls2009morph} and the test-set of the Animal-10N dataset \citep{song2019selfie} formed the test-set. The images were resized to 128 pixels in height and width, and were randomly flipped in the horizontal direction at train-time.   
\paragraph{}
We trained four CNN-models on this data. The first CNN-model and the second CNN-model have the same architecture, but use different losses. The first CNN used the margin-loss \citep{sabour2017dynamic} for its predictions. The second CNN used the cross-entropy loss. The architecture used for these two models uses regular correlations in its layers. Further, the architecture also used max-pooling. The third model is based on the Resnet architecture, and used the margin loss for its predictions. The fourth architecture used group-equivariant convolutions \citep{cohen2016group}, and used the margin loss for its predictions. The exact details of the models and their training can be found in our code which avaliable \url{https://github.com/compositionalcapsules/entropy_regularised_capsule}. We present the accuracy obtained from the models at the last epoch of training in Table \ref{cnnaccuracies}.
\begin{table}[h]
	\centering
	\caption{Accuracies of four CNN-models on the face vs. non-face task, averaged over three runs.} 
	\begin{tabular}{|p{1cm}||p{1cm}||p{1cm}|}
			\hline
			Method&Mean&Standard deviation\\
			\hline
			CNN1 & 100.00\%& 0.00\%\\
			\hline
			CNN2 & 100.00\%& 0.00\%\\
			\hline
			CNN3 & 99.93\%& 0.00\%\\
			\hline
			CNN4 & 100.00\%& 0.00\%\\
			\hline
		\end{tabular}\label{cnnaccuracies}
\end{table}
\paragraph{}
We further trained several capsule network models on this task. The capsule networks are based on two architectures, the specifications of which are available at \url{https://github.com/compositionalcapsules/entropy_regularised_capsule}. Each model is under a different amount of regularisation. They all use margin loss as the classification loss. The entropy-loss is added to this loss in a weighted-sum. The weight gives the amount of the regularisation. One unregularised model and a regularised according to a schedule, based on the first architecture are trained for this task. The schedule slowly increases the weight of the entropy-loss, while reducing the weight of the margin loss as the training continues. We term these models, unregcaps1 and schcaps1 to denote the unregularised model and the model regularised according to a schedule.

\begin{table}[h]
	\centering
	\caption{Accuracies and entropies of capsule network models on the face vs. non-face task, averaged over three runs.} 
	\resizebox{0.4\linewidth}{!}{\begin{tabular}{|p{1.5cm}|p{1.5cm}||p{1.5cm}||p{1.5cm}||p{1.5cm}||p{1.5cm}|}
		\hline
		Method&Mean of accuracy&Standard deviation of accuracy&Mean of entropy&Standard deviation of entropy\\
		\hline
		Unregcaps1& 99.96\%& 0.01\%& 2.29& 0.01\\
		\hline
		schcaps1& 99.44\%& 0.08\%& 0.004& 0.001\\
		\hline
		equalcaps1& 99.96\%& 0.00\%& 5.0& 5.0\\
		\hline
		sdacaps1& 99.98\%& 0.01\%& 2.73& 0.01\\
		\hline
		unregcaps2& 99.99\%& 0.01\%& 2.09& 0.05\\
		\hline
		0.4caps2& 96.63\%& 2.03\%& 0.012& 0.008\\
		\hline
		0.8caps2& 99.29\%& 0.59\%& 0.0002& 0.0001\\
		\hline
		schcaps2& 99.93\%& 0.07\%& 0.005& 0.004\\
		\hline
		equalcaps2& 99.99\%& 0.00\%& 5.0& 0.0\\
		\hline
		sdacaps2& 99.99\%& 0.00\%& 2.47& 0.00\\
		\hline 
	\end{tabular}}\label{capsuleresults}
\end{table}

\paragraph{}
For the second architecture, we train four models. These are an unregularised model, a model regularised by weighting the entropy-loss by 0.4 and the margin-loss by 0.6, a model regularised by weighting the entropy-loss by 0.8 and the margin-loss by 0.2 and a model regularised according to a schedule that increases the weight of the entropy-loss while reducing the weight of the margin-loss as the training continues. 
We use the terms unregcaps2, 0.4caps2, 0.8caps2, schcaps2 to denote these models.
\paragraph{}
We also trained capsule network models that give equal routing-weights to all the predictions, and models that uses the scaled-distance-agreement routing algorithm in \citep{peer2018increasing}. We use equalcaps1, equalcaps2, sdacaps1, and sdacaps2 to denote these models.
The code for the architectures and the training is given in our code. The accuracies and the mean of the entropy-loss for the test-set are given in Table \ref{capsuleresults}.
\paragraph{}
All the models, CNNs and capsule networks, performed very well in this task. However, 0.4caps2 is not as high-performing and has a large standard deviation in its results. It still is able to classify with above 96\% accuracy, and can be considered a good model for classification. We used the trained models in these tasks to detect changes in compositionality of the face images.
\paragraph{}
Specifically, we modify the face-images of the test-set by interchanging parts of the face. We interchange among the eyebrows, the eyes, the nose and the mouth regions. We do so by detecting the landmark points for the face, and then finding the regions based on this. The regions are resized while being changed as they could be of different sizes. The modified test-images thus are from face-images, but do not have the same compositionality.
\paragraph{}
A model that is compositional
will not be as activated on such images as much as they would be on faces. Thus, the mean activation for the predictions will be lower for compositional models on these images. The mean of the activations for the test-data is given in Table \ref{compostionalitydata}.
\begin{table}[h]
	\centering
	\caption{Mean activations of the predictions for face data} 
	\resizebox{0.3\linewidth}{!}{\begin{tabular}{|p{1.5cm}||p{1.5cm}||p{1.5cm}|} 
			\hline
			Model&Mean activation for faces& Mean activation for transformed faces\\
			\hline
			CNN1& 0.633& 0.633 \\
			\hline
			CNN2&1.00&0.984 \\
			\hline 
			CNN3&0.632&0.592 \\
			\hline
			CNN4&0.629&0.628 \\
			\hline
			Unregcaps1&0.730&0.646 \\
			\hline
			schcaps1& 0.727& 0.546 \\
			\hline
			equalcaps1& 0.727& 0.714 \\
			\hline 
	\end{tabular}\label{compostionalitydata}}
    \resizebox{0.3\linewidth}{!}{\begin{tabular}{|p{1.5cm}||p{1.5cm}||p{1.5cm}|}
                 \hline
                 Model& Mean activation for faces& Mean activation for transformed faces\\
                 \hline
                 sdacaps1& 0.730& 0.632 \\
                 \hline
                 unregcaps2& 0.730\%& 0.679 \\
                 \hline
                 0.4caps2& 0.714& 0.391 \\
                 \hline
                 0.8caps2& 0.725& 0.4836 \\
                 \hline
                 schcaps2& 0.729& 0.591\\
                 \hline
                 equalcaps2& 0.729& 0.719\\
                 \hline
                 sdacaps2& 0.730& 0.702\\
                 \hline
    \end{tabular}}
\end{table}
\paragraph{}
This experiment is an answer to papers \citep{paik2019capsule}, \citep{gu2020improving} that show that routing algorithms are not required  for performance. While deep learning models for computer vision do not need a routing-like mechanism for achieving high performance, we see that such models need not learn the compositionality of their inputs.
The CNN-models perform very poorly in this regard, showing no change in the activations. This is both in models that have max-pooling, and those which do not use it. Max-pooling is thought to reduce the ability of CNNs to learn spatial relationships due to the fact that the position of the original activation is not preserved, and this could cause a loss of spatial relationships among activations. We see that even when no pooling is used, CNNs do not detect learn compositional structures well.
\paragraph{}
We see that the performance of capsule networks, in detecting changes in compositionality, improves with a lower entropy among routing-weights. 
The capsule network model that gives equal weight to each routing-weight performs similarly to a CNN. We also note that the architecture of the model is also important to learning accurate compositional representations; the entropy among routing-weights is a measure of the tree-like nature of the representation, the learned derivation-rules depend on the architecture. Our experiments show that accuracy of a model in tasks is not proof of a good representation. For computer vision, one aspect that must be considered is the compositionality of the model.
\subsection{Experiments on classification} \label{EXP2}
The second experiment that we conduct is to test the classification performance of entropy-regularised models. We trained and tested three sets of models on SmallNorb, 
CIFAR10 and FashionMNIST.
\begin{table}[h]
	\centering
	\caption{Accuracies (in \%) and entropies of unregularised and regularised capsule network models on SmallNORB, averaged over three runs.} 
	\resizebox{0.7\linewidth}{!}{\begin{tabular}{|p{1.5cm}|p{1.5cm}||p{1.5cm}||p{1.5cm}||p{1.5cm}||p{1.5cm}|}
			\hline
			Method&Mean of accuracy&Standard deviation of accuracy&Mean of entropy-loss&Standard deviation of entropy-loss\\
			\hline
			unregcaps1& 92.88& 0.40& 1.43& 0.04\\
			\hline
			0.4caps1& 92.08& 1.51& 0.003& 0.002\\
			\hline
			0.8caps1& 93.48& 0.37& 2.8$\times10^{-5}$& 1.6$\times10^{-5}$\\
			\hline
			sch1caps1& 90.05& 0.53& 0.21& 0.04\\
			\hline
			sch2caps1& 90.28& 0.41& 0.18& 0.04\\
			\hline
			sdacaps1& 92.81& 0.31& 1.57& 0.06\\
			\hline 
	\end{tabular}}\label{capsuleaccuracySmallNORB}
	\resizebox{0.7\linewidth}{!}{\begin{tabular}{|p{1.5cm}|p{1.5cm}||p{1.5cm}||p{1.5cm}||p{1.5cm}||p{1.5cm}|}
			\hline
			Method&Mean of accuracy&Standard deviation of accuracy&Mean of entropy-loss&Standard deviation of entropy-loss\\
			\hline
			unregcaps2& 94.33& 0.39& 4.66& 0.02\\
			\hline
			0.4caps2& 94.29& 0.08& 0.004& 0.002\\
			\hline
			0.8caps2& 94.10& 0.72& 0.0003& 0.0002\\
			\hline
			sch1caps2& 93.85& 0.81& 0.39& 0.52\\
			\hline
			sch2caps2& 93.39& 0.63& 0.43& 0.19\\
			\hline
			sdacaps2& 63.09& 6.24& 4.55& 0.19\\
			\hline 
	\end{tabular}}
	\resizebox{0.7\linewidth}{!}{\begin{tabular}{|p{1.5cm}|p{1.5cm}||p{1.5cm}||p{1.5cm}||p{1.5cm}||p{1.5cm}|}
			\hline
			Method&Mean of accuracy&Standard deviation of accuracy&Mean of entropy-loss&Standard deviation of entropy-loss\\
			\hline
			unregcaps3& 91.40& 0.66& 1.36& 0.02\\
			\hline
			0.4caps3& 92.84& 0.20& 0.011& 0.003\\
			\hline
			0.8caps3& 93.24& 0.72& 0.001& 0.0002\\
			\hline
			sch1caps3& 89.12& 0.63& 0.25& 0.16\\
			\hline
		    sch2caps3& 89.98& 0.45& 0.14& 0.03\\
		    \hline
			sdacaps3& 90.87& 2.18& 1.44& 0.02\\
			\hline 
	\end{tabular}}
\end{table}
\begin{table}[h]
	\centering
	\caption{Accuracies and entropies of unregularised and regularised capsule network models on CIFAR10, averaged over three runs.} 
	\resizebox{0.7\linewidth}{!}{\begin{tabular}{|p{1.5cm}|p{1.5cm}||p{1.5cm}||p{1.5cm}||p{1.5cm}||p{1.5cm}|}
			\hline
			Method&Mean of accuracy&Standard deviation of accuracy&Mean of entropy-loss&Standard deviation of entropy-loss\\
			\hline
			unregcaps1& 91.48& 0.09& 3.47& 0.07\\
			\hline
			0.4caps1& 90.46& 0.29& 0.04& 0.009\\
			\hline
			0.8caps1& 90.31& 0.69& 0.01& 0.001\\
			\hline
			sch1caps1& 91.02& 0.05& 0.03& 0.008\\
			\hline
			sch2caps1& 91.45& 0.09& 0.02& 0.002\\
			\hline
			sdacaps1& 90.17& 0.17& 2.01& 0.02\\
			\hline 
	\end{tabular}}\label{capsuleaccuracyCIFAR10}
	\resizebox{0.7\linewidth}{!}{\begin{tabular}{|p{1.5cm}|p{1.5cm}||p{1.5cm}||p{1.5cm}||p{1.5cm}||p{1.5cm}|}
			\hline
			Method&Mean of accuracy&Standard deviation of accuracy&Mean of entropy-loss&Standard deviation of entropy-loss\\
			\hline
			unregcaps2& 91.09& 0.12& 8.98& 0.01\\
			\hline
			0.4caps2& 90.50& 0.27& 0.044& 0.0008\\
			\hline
			0.8caps2& 73.68& 0.31& 0.021& 0.00074\\
			\hline
			sch1caps2& 90.91& 0.08& 0.29& 0.0011\\
			\hline
			sch2caps2& 91.08& 0.12& 0.03& 0.004\\
			\hline
			sdacaps2& 89.23& 0.13& 5.19& 0.02\\
			\hline 
	\end{tabular}}
	\resizebox{0.7\linewidth}{!}{\begin{tabular}{|p{1.5cm}|p{1.5cm}||p{1.5cm}||p{1.5cm}||p{1.5cm}||p{1.5cm}|}
			\hline
			Method&Mean of accuracy&Standard deviation of accuracy&Mean of entropy-loss&Standard deviation of entropy-loss\\
			\hline
			unregcaps3& 93.65& 0.17& 1.87& 0.03\\
			\hline
			0.4caps3& 93.14& 0.16& 0.0042& 0.00079\\
			\hline
			0.8caps3& 92.37& 1.25& 0.00037& 0.00034\\
			\hline
			sch1caps3& 91.02& 0.05& 0.035& 0.008\\
			\hline
			sch2caps3& 92.99& 0.17& 0.16& 0.03\\
			\hline
			sdacaps3& 85.47& 0.37& 2.06& 0.005\\
			\hline 
	\end{tabular}}
\end{table}
\begin{table}[h]
	\centering
	\caption{Accuracies and entropies of unregularised and regularised capsule network models on FashionMNIST, averaged over three runs.} 
	\resizebox{0.7\linewidth}{!}{\begin{tabular}{|p{1.5cm}|p{1.5cm}||p{1.5cm}||p{1.5cm}||p{1.5cm}||p{1.5cm}|}
			\hline
			Method&Mean of accuracy&Standard deviation of accuracy&Mean of entropy-loss&Standard deviation of entropy-loss\\
			\hline
			unregcaps1& 92.73& 0.01& 1.05& 0.01\\
			\hline
			0.4caps1& 90.33& 0.09& 0.0003& 1.09$\times10^{-5}$\\
			\hline
			0.8caps1& 89.99& 0.98& 0.004& 0.0005\\
			\hline
			sch1caps1& 92.02& 0.21& 0.112& 0.008\\
			\hline
			sch2caps1& 92.10& 0.239& 0.159& 0.005\\
			\hline
			sdacaps1& 92.87& 0.049& 0.019& 0.002\\
			\hline 
	\end{tabular}}\label{capsuleaccuracyFashionMNIST}
	\resizebox{0.7\linewidth}{!}{\begin{tabular}{|p{1.5cm}|p{1.5cm}||p{1.5cm}||p{1.5cm}||p{1.5cm}||p{1.5cm}|}
			\hline
			Method&Mean of accuracy&Standard deviation of accuracy&Mean of entropy-loss&Standard deviation of entropy-loss\\
			\hline
			unregcaps2& 93.67& 0.09& 5.053& 0.021\\
			\hline
			0.4caps2& 92.93& 0.48& 0.03& 0.01\\
			\hline
			0.8caps2& 92.88& 0.73& 0.002& 0.002\\
			\hline
			sch1caps2& 92.81& 0.05& 0.49& 0.06\\
			\hline
			sch2caps2& 93.13& 0.11& 0.63& 0.06\\
			\hline
			sdacaps1& 93.32& 0.03& 5.26& 0.07\\
			\hline 
	\end{tabular}}
	\resizebox{0.7\linewidth}{!}{\begin{tabular}{|p{1.5cm}|p{1.5cm}||p{1.5cm}||p{1.5cm}||p{1.5cm}||p{1.5cm}|}
			\hline
			Method&Mean of accuracy&Standard deviation of accuracy&Mean of entropy-loss&Standard deviation of entropy-loss\\
			\hline
			unregcaps3& 93.05& 0.23& 1.48& 0.03\\
			\hline
			0.4caps3& 92.42& 0.25& 0.04& 0.003\\
			\hline
			0.8caps3& 92.39& 0.19& 0.005& 0.001\\
			\hline
			sch1caps3& 92.10& 0.020& 0.52& 0.25\\
			\hline
			sch2caps1& 92.30& 0.08& 0.62& 0.19\\
			\hline
			sdacaps3& 92.53& 0.19& 1.41& 0.00\\
			\hline 
	\end{tabular}}
\end{table}

Each model is trained under different combinations of the classification-loss and entropy-regularisation. The regularisation is done by a weighted-sum of the entropy-loss and the classification-loss. The weights are fixed or changed according to a schedule. We use a weight of 0.6 for the entropy-loss and 0.4 for the classification-loss in one case and 0.8 for the entropy-loss and 0.4 for the classification-loss in the other. We also train with one schedule which reduces the weight of the classification-loss and increases the weight of the entropy-loss as the training continues, and the second which keeps the classification-loss unweighted and increases the weight of the entropy-loss. 
We further train a model that uses the routing algorithm in \cite{peer2018increasing}.
The naming conventions of these models follow from section \ref{EXP1}. Note that the model used for CIFAR10 is different than the ones used in SmallNORB and CIFAR10.
\paragraph{}
The accuracy of the models is given in the Tables \ref{capsuleaccuracySmallNORB}, 
\ref{capsuleaccuracyCIFAR10} and  \ref{capsuleaccuracyFashionMNIST}. The mean entropy-loss for the test-set for the models is given along with the accuracies. In general, the entropy-regularised models perform similarily to the unregularised models and the models which use the routing algorithm in \cite{peer2018increasing}. One observation is that the results of entropy regularisation can lead to variations in accuracy across runs - reflecting the fact that optimising with a loss for every layer is not easy. However, our goal here is to show that under identical training conditions, models which are compositional perform on par with unregularised models in terms of accuracy. We did not aim for state-of-the-art results which may be possible with explicit hyperparameter tuning.  
%
\section{Conclusion}
We present a grammar-based framework for describing the compositionality of CNNs and capsule networks. We show that CNN-models do not have a compositional structure as they do not have a means of assigning importances to derivation-rules. We show that capsule networks with dynamic routing can activate derivation-rules based on the part-whole relations of the input.
\paragraph{}
We present an entropy-loss for improving the ability of capsule networks to detect compositional structures in data. We show, by experiments on face-data, that this entropy-loss improves the compositional-structure of capsule networks. Capsule networks that use the entropy-loss are able to better detect changes in compositional structure than unregularised capsule networks, while CNN-models perform poorly in this. We also see that capsule networks that use this entropy-loss perform similarily to unregularised capsule models.
\paragraph{}
In order to improve the performance of capsule networks whose entropy among routing-weights is low, we believe that routing algorithms that ensure a low entropy must be developed, as also methods to reduce the effect of background. Another area that could improve the performance of these models would be developing architectural units for better training.

\bibliography{iclr2021_conference}
\bibliographystyle{iclr2021_conference}
%

\end{document}